\documentclass[a4paper,twoside]{article}

\usepackage{multirow}
\usepackage{epsfig}
\usepackage{subcaption}
\usepackage{calc}
\usepackage{amssymb}
\usepackage{amstext}
\usepackage{amsmath}
\usepackage{amsthm}
\usepackage{multicol}
\usepackage{pslatex}
\usepackage{apalike}
\usepackage{algorithm2e}
\usepackage[bottom]{footmisc}
\usepackage{SCITEPRESS}     % Please add other packages that you may need BEFORE the SCITEPRESS.sty package.

\begin{document}

\title{Hate Speech Detection Using Cross-Platform Social \\Media Data In English and German Language}

\author{\authorname{Gautam Kishore Shahi\sup{1}\orcidAuthor{0000-0001-6168-0132}, Tim A. Majchrzak\sup{2}\orcidAuthor{0000-0003-2581-9285} }
\affiliation{\sup{1}University of Duisburg-Essen, Germany}
\affiliation{\sup{2}University of Agder, Norway}
\email{gautam.shahi@uni-due.de}
}

\keywords{Hate speech, YouTube, User comments, Cross-platform, Multilingual data}

\abstract{Hate speech has grown into a pervasive phenomenon, intensifying during times of crisis, elections, and social unrest. Multiple approaches have been developed to detect hate speech using artificial intelligence, but a generalized model is yet unaccomplished. The challenge for hate speech detection as text classification is the cost of obtaining high-quality training data. This study focuses on detecting bilingual hate speech in YouTube comments and measuring the impact of using additional data from other platforms in the performance of the classification model. We examine the value of additional training datasets from cross-platforms for improving the performance of classification models. We also included factors such as content similarity, definition similarity, and common hate words to measure the impact of datasets on performance. 
Our findings show that adding more similar datasets based on content similarity, hate words, and definitions improves the performance of classification models. The best performance was obtained by combining datasets from YouTube comments, Twitter, and Gab with an F1-score of 0.74 and 0.68 for English and German YouTube comments. }

%The performance of the BERT-based classification model was measured in different combinations of datasets. 

\onecolumn \maketitle \normalsize \setcounter{footnote}{0} \vfill

\section{\uppercase{Introduction}}
\label{sec:introduction}

Hate speech on social media has become a widespread problem on the Internet~\cite{jahan2023systematic}. With easy access to social media platforms, such as Twitter (now \emph{X}), YouTube, or Gab, the amount of hate speech has been increasing \cite{shahi2022mitigating} for decades. The topic of hate speech is linked to global developments and recent crises, such as hate speech on the Russia-Ukraine conflict \cite{di2023hate}, COVID-19 \cite{shahi2022mitigating}, and ongoing elections in different countries of the world. Different social media platforms have their own data formats and guidelines, allowing users to post content in various media formats, and languages. For example, on YouTube, users can post hate speech as videos or comments. 

Hate speech detection is mainly studied in English and for specific platforms such as Twitter \cite{siegel2020online}, Facebook \cite{del2017hate}, and YouTube \cite{doring2020gendered}. A hate speech classification model needs fine-grained annotated training data. Gathering high-quality training data is expensive and time-consuming \cite{shahi2022amused} for training machine learning models. Previous research shared hate speech datasets in different languages \cite{poletto2021resources,al2019detection,fortuna2018survey}. However, such data sets vary considerably by social media platform, annotation goal (such as offensive language, abusive language, hate speech), period of data collection, and other dimensions \cite{al2019detection}. There are notable differences in the linguistic style of comments posted on different platforms (such as adherence to standard spelling and grammar and usage of emojis). Hence, prior research and datasets are dissimilar and not readily generalizable, especially across different languages. 

With the advancement of Generative Artificial Intelligence, mainly the latest generative Artificial Intelligence (GAI) models for ChatGPT \cite{wullach2020towards}, such as data annotation data have been tested for hate speech. However, there is a need for ground truth to verify the annotation quality; annotation quality depends on the given definition, ethical implications, and local law \cite{li2023hot}. Hence, quality annotated data is still required for hate speech detection. 

Social media platforms are mandated in some jurisdictions to delete hateful messages within a certain time frame once they are reported as part of content moderation \cite{wang2023content}. As a result, thousands of posts need to be checked automatically for whether or not they contain hateful content \cite{bayer2020hate}. 
This study proposes a hate speech detection model for bilingual YouTube comments. User comments were collected from YouTube videos in English and German and covered different social topics for instance politics, LGBT rights, and immigration. Comments are annotated for hate speech. To solve the problem of data annotation on a large scale, we collected and annotated a small number of YouTube comments and explored our approach by reusing existing datasets to enhance the performance of the classifier. We propose the following research question (RQ):
%This work focuses on exploring the performance of the classification model for YouTube comments with additional data from external sources. 

\emph{How does the performance of the Classifier for hate speech detection change by adding additional training data?}

To answer the RQ, we collected users' comments from YouTube videos. We annotated  1,892 English and 6,060 German comments from 190 YouTube videos on different topics. Along with that, we used eight existing datasets from different platforms in German and English. First, we compute the \textit{dataset similarity} based on definition, text, and hate words. In similarity measures, text and hate speech provide content similarity from datasets and definition similarity provides choice of hate speech. Further, we systematically compare the predictive performance of hate speech classifiers when the training data is augmented with various kinds of external data across different platforms in two different settings. We employed \textit{cross-data set training}, i.e. augmenting the training data with an external data set (from the same social media platform and in the same language). Finally, we used \textit{cross platform training}, i.e. augmenting the training data with external data from a different social media platform. 

For evaluation, we considered the performance of the machine learning model in terms of precision, recall, and F1 score, the similarity of hate words from the dataset, and the definition of hate speech in the dataset. Apart from the performance of the classification model, the definition and hate word similarity provide an overview of linguistic content.
The key contribution of this work is twofold. First, we provide a dataset for hate speech on YouTube comments in English and German.
Second, we explore existing hate speech datasets to enhance the performance of the hate speech classification model.

This article is organized as follows. We discuss related work in Section~\ref{sec:2} followed by the research method in Section~\ref{sec:3}. We then show our experiment and results in Section~\ref{sec:4}, before discussing the findings in Section~\ref{sec:5}. Finally, we discuss the ideas for future work in Section~\ref{sec:6}. 

\section{\uppercase{Related Work and Background}}
\label{sec:2}
Machine learning is the dominant approach to text classification in various domains such as political sentiment analysis \cite{rochert2020opinion}, detecting incivility and impoliteness in online discussions \cite{stoll2020detecting} as well as the classification of political tweets \cite{charalampakis2016comparison}. Especially the state-of-the-art technique BERT (Deep Bidirectional Transformers for Language Understanding) \cite{malmasi2017detecting} has used for the detection of hate speech \cite{salminen2020developing,aggarwal2019ltl,liu2019nuli,zampieri2019semeval}. The text classifier aims to train a robust classifier to recognize hate comments globally, i.e., on different platforms and languages. For the identification and training of the models, several studies classify hate speech using deep neural network architectures or standard machine learning algorithms. Wei et al. compare machine learning models on different public datasets \cite{wei2017convolution}. The results indicate that their approach of a convolution neural network outperforms the previous state-of-the-art models in most cases \cite{wei2017convolution}. A recent study used transfer learning \cite{yuan2023transfer} to train 37,520 English tweets, showing a trend towards more complex models and better results.
Prior studies have analyzed hate speech on YouTube; one study highlights the hate speech on YouTube on Syrian refugees \cite{aslan2017online}.
In another study, hate speech on gender is studied on a small dataset quantitatively \cite{doring2019fail}. 

Previous studies have mainly focused on detecting hate speech in different languages \cite{ousidhoum2019multilingual} and on comparing different social media platforms \cite{salminen2020developing}. Considering the past studies about hate speech, it is noticeable that many studies only concentrate on one platform or specific language, such as English \cite{waseem2016you}, German \cite{ross2016hatespeech}, Spanish~\cite{ben2016hate} and Italian \cite{del2017hate} to train a machine learning model to automatically classify and predict which unseen texts can be considered hate speech.
Ousidhoum et al. applied a multilingual and multitasking approach to train a classifier on three languages (English, French, and Arabic) based on Twitter tweets using a comparison of traditional machine learning and deep learning models \cite{ousidhoum2019multilingual}. 

Previous research uses human-annotated datasets for supervised hate speech detection. Besides the different methods for training the model, different code books are used for data annotation. At the same time, \cite{davidson2017automated,waseem2016hateful} opted for a multi-label procedure (hateful, offensive (but not hateful), and neither \cite{davidson2017automated}; racist, sexist \cite{waseem2016hateful}), the annotation of the data of \cite{ross2016hatespeech} was collected with a binary labeling schema (hate speech as yes or no). 

However, research is needed considering the classifier's performance in multilingual contexts in combination with datasets from social media platforms. Salminen et al. \cite{salminen2020developing} points out that the mono-platform focus in hate speech detection research is problematic, as there are no guarantees that the models developed by researchers will generalize well across platforms. This issue underscores the importance of our research in exploring cross-platform generalization. 

Fortuna et al. have shown that merging and combining different datasets can enhance the overall performance of classification models \cite{fortuna2018survey}. However, a more systematic evaluation is needed to determine the extent of improvement when a classifier is trained on one dataset and then another is added. This approach could potentially lead to significant advancements in hate speech detection. 

The role of generative AI models such as chatGPT in generated annotated data without any background truth is still under exploration \cite{li2023hot}. Consequently, hate speech detection still depends on gathering human-annotated data to build a classification model.

\section{\uppercase{Research Method}}
\label{sec:3}
\begin{figure*}[!htbp]
\centering
\includegraphics[width=0.6\textwidth]{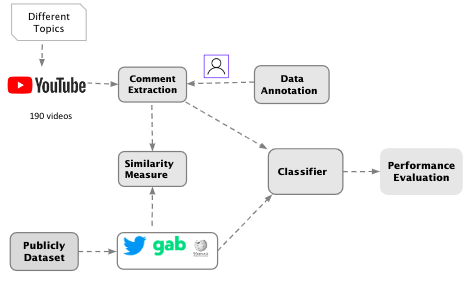}
\caption{Methodolodoy used for the hate speech detection}
\label{fig:method}
\end{figure*}

This section explains the approach used for building the classification model in the study, i.e., data collection, data annotation, similarity measurement, the classification approach, and the evaluation strategy. The research method is depicted in Figure~\ref{fig:method}.

\subsection{Data Collection}
First, we collected datasets in two steps: 1) Collecting YouTube comments in English and German. 2) Compiling publicly available data published in prior research on Twitter, Wikipedia, and Gab.

YouTube is a video-sharing platform where videos from across the world are uploaded. We used a self-developed Python crawler using pytube\footnote{https://pytube.io/en/latest/} (a Python library) and searched keywords on YouTube to gather video and their comments \cite{rochert2021networked}. To maintain the diversity in the dataset, we chose different controversial topics; such as \textit{politics, LGBT, immigration, abortion, war, sex education, entertainment, and sports}, the probability of having hate comment is more on videos on these topics. A total of 49,074 comments were extracted from 101 English YouTube videos and 89 German videos in multiple languages. Furthermore, we identified the language of comments and filtered German and English comments using fastText \cite{joulin2016fasttext}. Overall, we got 30,663 and 18,441 English and German comments, respectively, which are further filtered for data annotation, further explained in section \ref{sec:da}.

For external datasets, the criteria were to get publicly accessible datasets and use binary class for hate speech detection. We considered ten datasets from three different platforms (Twitter, Gab, Wikipedia) and provided a three-letter ID to each of them; the first letter indicates the language, the second the platform, and the third the number. For example, \emph{ET1} is the 1st English Twitter dataset. Overall, We used ten datasets; eight datasets were collected from external sources, and two datasets (EY1 and GY1) were collected and annotated by us. Table~\ref{data:dataoverview} gives an overview of the collected dataset. For each dataset, we provide the language, platform name, size of data, and percentage of hate speech annotated in the dataset.

\begin{table*}[tb!]
 \centering
    \label{qualifications2}
        \caption{Datasets compiled from different platforms in German and English}
    \begin{tabular}{|l|l|l|r|r|l|}
\hline
        \textbf{Data set} &  \textbf{Language} & \textbf{Platform}  & \textbf{Size}   & \textbf{\% Hate speech}   & \textbf{Availability}  \\
        \hline 
      
     EY1 & English & YouTube & 1,892 & 37.36 & Partial   \\ 
   ET1 \cite{mandl2019overview} & English &   Twitter   & 7,005 & 16.31 & Partial   \\ 
     ET2 \cite{davidson2017automated} & English &    Twitter   & 24,783 & 5.77 & Open     \\
      ET3 \cite{basile2019semeval} & English &  Twitter  & 12,971 & 42.10 & Partial  \\
     ET4 \cite{waseem2016hateful} & English & Twitter & 10,498 & 27.35 & Partial   \\ 
   EW1 \cite{jigsawchallenge} & English &   Wikipedia & 312,735 & 6.8~ & Open   \\
     EG1 \cite{gab2020pushshift} & English &    Gab  & 27,265 & 8.4~ & Open     \\ 
     \hline
      GY1  & German &  YouTube  & 6,060 & 4.6~ & Partial  \\ 
       GT1 \cite{mandl2019overview} & German & Twitter  & 4,469 & 2.3~ &  Partial   \\
   GT2 \cite{ross2016hatespeech}  & German &   Twitter   & 470 & 11.7~ & Partial    \\  \hline
    \end{tabular}
    \label{data:dataoverview}
\end{table*}

\subsection{Data Annotation}
\label{sec:da}

Each social media platform, community, organization, and government defines hate speech differently \cite{fortuna2018survey}. Before developing our codebook, several definitions of hate speech were studied, such as given by the EU code of conduct \cite{wigand2017speech}, ILGA \cite{ilga}, Nobata \cite{nobata2016abusive}, Facebook \cite{facebook}, YouTube \cite{youtube} and Twitter \cite{twitter}. 

Multiple definitions of hate speech have been proposed. Nockleby defines \textit{hate speech as any communication that disparages a person or a group based on some characteristics such as race, color, ethnicity, gender, sexual orientation, nationality, religion, or other characteristics} \cite{nockleby2000hate}. Fortuna \& Nunes discuss the definition of hate speech, and the criteria to remove it varies across each social media platform \cite{fortuna2018survey}, ILGA Europe \cite{ilga} and code of conduct between European Union and companies \cite{euc}. 

%For data annotation, we have analyzed different definitions of hate speech. 
We found a lack of generic definitions that apply to cross-platform and different languages. We, thus, define hate speech as: \textit{Language that expresses prejudice against a person from a group (e.g. Barack Obama for being black) or a particular group (such as black people), primarily based on ethnicity, religion, or sexual orientation.
The message should be about a specific group based on skin color (e.g. black, white), religion (e.g. Hindu, Muslim, Christian), gender (e.g. male, female, non-binary), class (e.g. rich, poor), ethnicity (e.g. Asian, European), sexual orientation (e.g. lesbian, gay, bisexual, transgender), nationality (e.g. Indian, German), physical appearance (e.g. beautiful, ugly, short, tall), disability (e.g. handicapped) or disease (e.g. ill or fit).} 

The proposed definition was used to annotate hate speech in YouTube comments. Annotators were provided with different sets of examples of hate speech and non-hate speech in both languages. Three annotators with a bachelor degree in computer science with backgrounds in both languages carried out the annotation process.  %TODO I think they need to be mentioned in the annotations, which for now should be blinded.
%do I need to do anything here?
%TODO I suggest we add a section Acknowledements at the end of the document but insert that "Acknowledements have been blinded for review". In the camera ready version those that helped with coding need to be mentioned.
%ok
We conducted a test annotation for both languages and found that the number of hate speech comments is less for German, so we randomly filtered 2000 and 6,500 comments for English and German.
While annotating, we also identify the hateful words mentioned in the comment. 
The majority vote out of three annotators is considered for the final label. To measure the reliability among annotators, Cohen's kappa \cite{mchugh2012interrater} is used for intercoder reliability. We got 1,892 English and 6,060 German YouTube comments, resulting in Cohen's kappa of 0.86, indicating acceptable interrater reliability. 

\subsection{Similarity Measurement}
%We aim to measure the existing datasets' role in improving the classifier's performance. 
We collected eight different datasets for evaluating cross-platform bilingual classification models. Each dataset was collected differently on various controversial topics, so we computed the similarity among datasets based on definition, hate words, and content as described below. The similarity measures are decided based on the factors affecting the dataset, such as the definition mentioned in the codebook, textual content, and the amount of hate words is content and region-dependent.

\textbf{Definition Similarity} Each dataset was annotated based on a different codebook (extracted from a scientific publication mentioning the dataset), so we measured the similarity of the definition of hate speech on each dataset. The annotation codebook significantly affects the quality of annotated data. For the definition similarity, we conducted an online study using Prolific \cite{prolific} as discussed in Section~\ref{sec:4}.

\textbf{Content Similarity} We measured the similarity between the content among two datasets. Content similarity indicates common words mentioned in a different dataset. We used a sentence-transformer model for computing the content similarity \cite{kazemi2022research}. We converted text from each dataset into vectors and simultaneously computed the cosine similarity between the two datasets. The content similarity measures the semantic similarity of datasets. 

\textbf{Hate word Similarity} In hate speech detection, the text contains hate words referring to a person or group of people. First, we collected the hate words from literature and added hate words from YouTube comments defined by annotators. Using these hate words, we filtered hate words from each dataset, computed the intersection of hate words between two data sets, and divided it by the total number of hate words in the two datasets. It helps to analyze the similarity of the presence of hateful content within two datasets.

The definition similarity score was calculated using user study, basic details were provided to participants, and they were asked to vote for similarity on a 10-point scale. Finally, we normalized the average value using the formula (n-1)/9, where n is the average of the similarity score to show the similarity score for the definition. For content similarity, we used sentence embeddings of text as discussed by \cite{kazemi2022research}; for hate words, we used text matching using Python. A detailed description of all three similarities is discussed in Section 4.

\subsection{Classification Model}
We followed the traditional natural language processing (NLP) data cleaning technique method using the Natural Language Toolkit (NLTK) library \cite{loper2002nltk}. This includes removing short words (i.e., less than three characters), emails, and hyperlinks.

We used different machine models for the classification model, such as the Support Vector Machine (SVM), Long short-term memory (LSTM), Logistic Regression, and the state-of-the-art Bidirectional encoder representations from transformers (BERT) model. First, we used the annotated YouTube comments and measured performance. Later, we added the dataset in different combinations (from same and cross-platform) to the YouTube comments and measured the performance of the classification model. 

We evaluated the performance of the classification model generated model in terms of precision, recall, and F1. We provided the model evaluation for both positive and negative classes.

\section{\uppercase{Experiment and Results}}
\label{sec:4}
We computed the similarity of datasets based on definition, content, and hate words. All three measures indicate the similarity of the different datasets concerning YouTube comments. The goal is to find the relationship between these similarities and classification performance. The result obtained from similarity measures is shown in Tables~\ref{similarityEN1}, \ref{similarityEN2}, and~\ref{similarityEN3} for English, as well as ~\ref{similarityDE} for German.

\textit{Definition similarity} We conducted an online survey with 100 participants on Prolific. The participants were from all over the globe, with an average age of 26 years: 60\%  male and 40\% female, 66\% working professionals, and the remaining students.  Out of eight external datasets, only six defined hate speech explicitly. A combination of two definitions from six datasets, 15 combinations of definitions of similarity, was provided to participants, and they were asked to vote for similarity on a 10-point scale. Average votes from 90 participants (10 participants were excluded because their response time was \emph{too} fast, indicating they voted without really pondering about it) were calculated to measure the similarity for each similarity. Finally, we normalized the average value using the formula (n-1)/9, where n is the average of the similarity score to show the similarity of the definition used in the data annotation.
The result shows that the definition used in English YouTube comments is similar to the definition of ET4, followed by ET5.

For \textit{Hate word similarity}, we filtered the hate words mentioned in our datasets using Hurtlex \cite{bassignana2018hurtlex}. We also screened the hate words mentioned in other datasets. We computed common hate words using text matching and represented the hate word similarity in percent. Based on the hate word similarity, most datasets share similar hate word datasets, and English Wikipedia contains words that are more similar to EY1. However, EG1 contains the maximum number of similar hate words in English, and GT2 is more similar to all datasets. So, for English, Gab has more hateful content, which is not true for German.

For the \textit{Content similarity}, we converted each dataset into a vector using the XLM-RoBERTa sentence transformer model \cite{reimers2020making} and computed the cosine similarity of each pair of datasets. The content similarity indicates the overall common words present in different datasets concerning our dataset. After computing the content similarity, EG1 has maximum similarity to EY1 for English. For German, GT1 has a definition that is more similar to GY1.

%After computing the content similarity, ET3 shares a definition that is more similar to EY1 for English. However, the definition used in EG1 was similar to all datasets.For German, GT1 has a definition that is more similar to EG1.

\begin{table*}[!tb]
\centering
\caption{Content similarity measures for the English dataset}
        \begin{tabular}{|c|c|c|c|c|c|c|c|}
 \hline
       & \textbf{EY1}  &   \textbf{ET1} &   \textbf{ET2} &   \textbf{ET3} &    \textbf{ET4} & \textbf{EW1} &  \textbf{EG1} \\
      \textbf{EY1}  & 1.00 & 0.31 & 0.29 & 0.36 & 0.44 & 0.47 & \textbf{0.56}\\
   \textbf{ET1} &   & 1.00 & 0.15 & 0.77 & 0.28 & 0.24 & 0.73\\
    \textbf{ET2} &   &  & 1.00 & 0.49 & 0.43 & 0.30 & 0.35 \\
       \textbf{ET3} &  &  &  & 1.00 & 0.38 & 0.30 & 0.72 \\
         \textbf{ET4} &  &  &  &  & 1.00 & 0.53 & 0.63\\
          \textbf{EW1} &  &  &  &  &   & 1.00 & 0.54\\
             \textbf{EG1} &  &  &  &  &  &  & 1.00\\  \hline
    \end{tabular}
\label{similarityEN1}
\end{table*}

\begin{table*}[!tb]
\centering
\caption{Hate word similarity measures for the English dataset}
\begin{tabular}{|c|c|c|c|c|c|c|c|}
\hline
       & \textbf{EY1}  &   \textbf{ET1} &   \textbf{ET2} &   \textbf{ET3} &    \textbf{ET4} & \textbf{EW1} &  \textbf{EG1} \\
      \hline
      \textbf{EY1}  & 1.00 & 0.72 & 0.88 & 0.86 & 0.78 & \textbf{0.92} & 0.91\\
   \textbf{ET1} &   & 1.00 & 0.71 & 0.75 & 0.73 & 0.77 & 0.80\\
    \textbf{ET2} &   &  & 1.00 & 0.83 & 0.84 & 0.89 & 0.88 \\
       \textbf{ET3} &  &  &  & 1.00 & 0.83 & 0.89 & 0.90 \\
         \textbf{ET4} &  &  &  &  & 1.00 & 0.83 & 0.84 \\
          \textbf{EW1} &  &  &  &  &   & 1.00 & 0.92\\
             \textbf{EG1} &  &  &  &  &  &  & 1.00\\   \hline
    \end{tabular}
\label{similarityEN2}
\end{table*}

\begin{table*}[!tb]
\centering
\caption{Definition similarity measures for the English dataset}
    \begin{tabular}{|c|c|c|c|c|c|c|c|}
    \hline
        & \textbf{EY1}  &   \textbf{ET1} &   \textbf{ET2} &   \textbf{ET3} &    \textbf{ET4} & \textbf{EW1} &  \textbf{EG1} \\
      \textbf{EY1}    & X	& 0.66 &	0.63&	\textbf{0.76} &	0.70 &	X  &	0.56 \\
   \textbf{ET1}   & & X & 0.64 &	0.72 &	0.62 &	X  &	0.57 \\
    \textbf{ET2} &   &  & X &	0.64 &	0.63 &	X &	0.63  \\
       \textbf{ET3} &  &  &  & X &	0.61 &	X & 0.59 \\
         \textbf{ET4} &  &  &  &  & X &	X &	0.59  \\
          \textbf{EW1} &  &  &  &  &   & X &	X	\\
             \textbf{EG1} &  &  &  &  &  &  & X \\  \hline
    \end{tabular}
\label{similarityEN3}
\end{table*}

\begin{table*}[!tb]
\caption{Similarity measures for the German dataset}
\begin{minipage}[t]{0.34\textwidth} \centering
  $a)$ $Content$ $Similarity$ \\
    \begin{tabular}{ |c|c|c|c| } 
  \hline
      & \textbf{GY}  &   \textbf{GT1} &   \textbf{GT2}  \\ \hline
      \textbf{GY}  & 1.00 &  \textbf{0.14}  & 0.10\\
   \textbf{GT1} &   & 1.00 & 0.11 \\
    \textbf{GT2} &   &  & 1.00  \\ \hline
       
    \end{tabular}
\end{minipage}
\begin{minipage}[t]{0.30\textwidth} \centering
  $b)$ $Hate word$ $Similarity$ \\
    \begin{tabular}{ |c|c|c|c| }   
 \hline
      & \textbf{GY}  &   \textbf{GT1} &   \textbf{GT2}  \\ \hline
      \textbf{GY}  & 1.00  & \textbf{0.35}  & 0.24\\
   \textbf{GT1} &   & 1.00 & 0.10 \\
    \textbf{GT2} &   &  & 1.00   \\ \hline
    \end{tabular}
\end{minipage}
\begin{minipage}[t]{0.30\textwidth} \centering
     $c)$ $Definition$ $Similarity$  \\ 

\begin{tabular}{ |c|c|c|c| } 
  \hline
      & \textbf{GY}  &   \textbf{GT1} &   \textbf{GT2}  \\ \hline
\textbf{GY}  & X & \textbf{0.56}  & 0.44\\
   \textbf{GT1} &   & X  & 0.54 \\
    \textbf{GT2} &   &  & X   \\ \hline
    \end{tabular}
\end{minipage}

\label{similarityDE}
\end{table*}

To implement classification models for traditional machine learning models, SVM has been implemented using the scikit-learn library \cite{scikit-learn}. For LSTM, we applied Text-to-Sequences to vectorize the data; for CNN, a Document-Term Matrix was used. We used Hugging Face's pre-trained distilbert-base-uncased~\cite{DistilBERT} model for the BERT model. The BERT model outperforms others, so the result and training procedures describe the BERT model.

For training, we first used EY1 for English and GY1 for German; we used 70\% of the corpus for training and the remaining 30\% for testing. The test dataset was kept separate for each dataset combination for evaluation. First, we trained the classifier for EY1 and GY1 and tested the performance separately, then for further training, we added different datasets along with the training set of EY1 and GY1 to keep the test set the same for evaluation. The class distribution of each dataset except ET4 and EG1 was highly imbalanced; most of the collected datasets needed to be balanced. We used undersampling with equal class distribution and the overall dataset for implementation because it gives better performance than the original dataset. We noted the number of hate classes and randomly filtered the same number from non-hate classes. 

We used a pre-trained BERT model with fine-tuning as a learning rate $3^{-5}$, batch size of 32, and sequence length of 128 to six epochs. We report the precision, recall, and F1 scores in both classes. In Table~\ref{tab:result}, we present the result obtained from the classification model on different combinations of datasets: Adding more datasets helps in increasing precision, recall, and F1 score. We emphasized the maximum score obtained for the evaluation measures as bold in the Table~\ref{tab:result}. The Table presents results from the best combinations of datasets for the BERT model only.

\begin{table*}[!httb] 
\centering
\caption{Summary of the Precision, Recall, F1 Score of classification model}
    \begin{tabular}{|c|c|c|c|c|c|c|}           \hline 
 \multirow{2}{*}{Target Dataset} &   \multirow{2}{*}{Trainig Datasets}    & \multicolumn{2}{c|}{Non-Hate}    &\multicolumn{2}{c|}{Hate} & \multirow{2}{*}{F1}               \\   \cline{3-6}
  & &   {Prec.} & {Rec.} & {Prec.}  & {Rec.}    &          \\   \hline 
 EY1 & EY1  & 0.67 & 0.55 & 0.62   & 0.73  & 0.64  \\   
 EY1  & EY1+ET1 & 0.72 & 0.14 & 0.52   & \textbf{0.95}  & 0.51            \\  
EY1 & EY1+ET2 & 0.61 & 0.87 & 0.77   & 0.43  & 0.64         \\   
 EY1 & EY1+ET3 & 0.60 & 0.93 & \textbf{0.84}   & 0.38  & 0.62               \\     
 EY1 & EY1+ET4 & 0.79 & 0.67 & 0.71   & 0.82  & 0.74  \\   
 EY1  & EY1+EW1 & 0.55 & 0.86 & 0.68   & 0.30  & 0.54             \\  
EY1 &  EY1+EG1 & 0.80 & 0.57 & 0.67   & 0.85  & 0.71           \\   
 EY1 & EY1+ET1+ET4 & 0.70 & 0.73 & 0.72   & 0.69  & 0.71               \\    
 EY1 & EY1+ET2+ET4 & 0.71 & 0.66 & 0.69   & 0.73  & 0.70  \\   
 EY1 & EY1+ET3+ET4 & 0.54 & 0.88 & 0.68   & 0.26  & 0.52  \\
 EY1  & EY1+ET4+EW1 & 0.81 & 0.47 & 0.63   & 0.89  & 0.66             \\  
EY1 & \textbf{EY1+ET4+EG1}  & \textbf{0.83} & \textbf{0.61} & \textbf{0.69}   & \textbf{0.87}  & \textbf{0.74}            \\   \hline
 GY1 &  GY1 & 0.77 & 0.60 & 0.67   & 0.82  & 0.71   \\  
  GY1 &  GY1+GT1 & 0.82 & 0.38 & 0.59   & \textbf{0.92}  & 0.62  \\
   GY1 &  GY1+GT1+GT2 & 0.72 & 0.60 & \textbf{0.66}   & 0.76  & 0.68  
   \\     \hline 
 \end{tabular}
 \label{tab:result}
 \end{table*}

\section{\uppercase{Discussion}}
\label{sec:5}
%One of the key challenges in building a machine learning model for hate speech detection is gathering labeled data. 
Collecting the data set from the social media platform and annotating the data is time-consuming. To overcome the problem, we collected and annotated YouTube comments built a classification model, and further used the existing dataset for training. To measure the role of the additional datasets in the performance of the classifier, we looked at literature works for finding datasets. However, only a few research share datasets as open-access
%many of them did not publish their data. 
Even when it is partially available (for instance, only the tweet IDs in Twitter (\emph{X}) data sets), the user needs to write a program to collect the data or have access to an existing one and know how to use it. With time, many social media posts are deleted by either social media platforms or users (cf. e.g. \cite{waseem2016you}). We could crawl only 10,498 out of 16,914 labeled entries. Due to platform restrictions from YouTube, prior work did not share users' comments publicly which makes it difficult to explore the performance of the classifier by adding users' comments from other hate speech. A publicly accessible dataset will ease the workload to detect hate speech on social media. 

During content similarity, we observed that a dataset that contains hate speech on general topics such as sexism, racism, targeted populations, vulgarity, and framing helps to improve the performance of the classification model. We found that \textit{Class imbalance} is a problem for hate speech detection; so far, all data sets are imbalanced. We observed that under-sampling data with an equal distribution of classes gives better results.  

%ET4 and EG1 improve the overall performance due to similar dataset topics.

We have evaluated the role of additional datasets for the performance of the classifier on different criteria. For English datasets, based on the similarity measure, \textit{Definition Similarity} often improves the classifier performance. For example, ET3 is very similar to EY1, but EG1 is dissimilar, and it still improves classification performance as a precision of hate class.
There is no direct relation in the performance model that a dataset with a similar definition gives the highest precision. \textit{Content Similarity} is directly related to enhancing the performance of the classification model; ET4 from the same platform improves the classification model, while the EG1 from cross-platform has the maximum similarity with EY1, which enhances the overall performance of the mode. So, there is a direct relation; the similar content improves the performance. \textit{Hate word Similarity} also does not help to conclude that EW1 does not improve the performance but hate word similarity with content similarity, and the performance increases such as EG1 increases the performance.

However, for German, based on the similarity in content, the performance of the classifier improved by adding GT1. The same trend follows for the hate word similarity and definition similarity. However, the similarity between datasets regarding content and hate words is less than that of English. 

Therefore, if the dataset is similar in terms of similarity measures, additional data improves the performance of the classification model. Overall, data annotation is a key challenge, but combining datasets having a similar issue with similar content and hate words improves the classification performance. A heuristic combination of a balanced dataset could improve precision, recall, and f1-score. There is no direct relation in increasing the dataset from the same platform, but cross-platform datasets improve performance such as ET4 and EG1 jointly improve recall and F1-score. In this study, the classification model was based on the traditional machine learning approach, however, the performance might change by applying a Large Language model such as ChatGPT. LLMs, pre-trained on vast amounts of data from the web which might capture deeper contextual and semantic interpretation of the text. 
By fine-tuning an LLM on the specific task, the classification model could better handle complexities which could potentially lead to better performance compared to traditional methods.

%TODO I tried to craft a bit of a philosophical ending to the discussion. Please edit as you deem fit.
Improvements to hate speech detection will ultimately have organizational and social consequences. If detecting hate speech is easier, there will be less room for social media platforms to not remove hateful content. The consequences this will have on social media users -- those affected by hate speech, those being ``bystanders'', and those posting hate speech -- will need to be targeted by further research. The consequences of potentially many posts being deleted are not clear, either. However, neither removal nor detection changes much of the roots of hate speech surfacing on social media.
While the work we present here is mainly technological in nature, countering hate speech is a truly interdisciplinary approach and will require cross-disciplinary efforts.

\section{\uppercase{Conclusion and Future Work}}
\label{sec:6}
In this study, we presented hate speech detection on YouTube comments and explored the classification model by adding external data from the same platform or cross-platform in English and German. We provided datasets with 1,892 and 6,060 English and German human-annotated. We tested the combination of different data sets and measured performance in terms of precision, recall, and F1 score. Along with the performance metric, we have used similarity measures to conclude the role of the external datasets in enhancing classifier performance. 

To enhance the model performance, we used external datasets from different platforms. We computed the similarity of datasets based on definition, content, and hate words and explored their importance for the classification models. Overall, we succeeded in implementing the performance of the model by adding an existing hate speech dataset and different patterns for both languages.

The practical application of this approach is to create a generalized classification model for hate speech detection over bilingual cross-platform. Our approach can be tested with a heuristic dataset combination in future work.  One of the further extensions of our work is to extend to use of additional datasets in different languages, such as Spanish and Italian. Another extension could be multi-class hate speech classification with classes such as hate speech, offensive, or profane.

\section*{Data Sharing} 
We have conducted all experiments on a macro level following
strict data access, storage, and auditing procedures for the sake of
accountability. Following the guidelines of the YouTube data-sharing policy, we will release comment IDs, YouTube video IDs, and a replication package to download the data. We share the link to the existing dataset for the data collected from another source at GitHub\footnote{https://github.com/Gautamshahi/BilingualYouTubeHateSpeech}. 
%A sample data set is provided at the time of review\footnote{https://tinyurl.com/44xzfdr9}.
%yes we can release the data, would it sample of it work?.
%TODO The sample includes the names of the posters, some of which appear to be real names. Is this all right from an ethical standpoint? This could be considered as revealing those people
%nice catch, we can remove 1st column as id, ok?

\section*{\uppercase{Acknowledgements}} We thank our annotators, Alexander Kocur, and Jessica Priesmeyer for completing the annotating task. 

\bibliographystyle{apalike}
{\small
\bibliography{example}}

\begin{thebibliography}{}

\bibitem[Aggarwal et~al., 2019]{aggarwal2019ltl}
Aggarwal, P., Horsmann, T., Wojatzki, M., and Zesch, T. (2019).
\newblock Ltl-ude at semeval-2019 task 6: Bert and two-vote classification for categorizing offensiveness.
\newblock In {\em Proceedings of the 13th International Workshop on Semantic Evaluation}, pages 678--682.

\bibitem[AI, 2018]{jigsawchallenge}
AI, J. (2018).
\newblock {\em Toxic Comment Classification Challenge}.

\bibitem[Al-Hassan and Al-Dossari, 2019]{al2019detection}
Al-Hassan, A. and Al-Dossari, H. (2019).
\newblock Detection of hate speech in social networks: a survey on multilingual corpus.
\newblock In {\em 6th International Conference on Computer Science and Information Technology}.

\bibitem[Aslan, 2017]{aslan2017online}
Aslan, A. (2017).
\newblock Online hate discourse: A study on hatred speech directed against syrian refugees on youtube.
\newblock {\em Journal of Media Critiques}, 3(12):227--256.

\bibitem[Basile et~al., 2019]{basile2019semeval}
Basile, V., Bosco, C., Fersini, E., Nozza, D., Patti, V., Pardo, F. M.~R., Rosso, P., and Sanguinetti, M. (2019).
\newblock Semeval-2019 task 5: Multilingual detection of hate speech against immigrants and women in twitter.
\newblock In {\em Proceedings of the 13th International Workshop on Semantic Evaluation}, pages 54--63.

\bibitem[Bassignana et~al., 2018]{bassignana2018hurtlex}
Bassignana, E., Basile, V., and Patti, V. (2018).
\newblock Hurtlex: A multilingual lexicon of words to hurt.
\newblock In {\em 5th Italian Conference on Computational Linguistics, CLiC-it 2018}, volume 2253, pages 1--6. CEUR-WS.

\bibitem[Bayer and Petra, 2020]{bayer2020hate}
Bayer, J. and Petra, B. (2020).
\newblock Hate speech and hate crime in the eu and the evaluation of online content regulation approaches.
\newblock EPRS: European Parliamentary Research Service.

\bibitem[Ben-David and Matamoros-Fern{\'a}ndez, 2016]{ben2016hate}
Ben-David, A. and Matamoros-Fern{\'a}ndez, A. (2016).
\newblock Hate speech and covert discrimination on social media: Monitoring the facebook pages of extreme-right political parties in spain.
\newblock {\em International Journal of Communication}, 10:1167--1193.

\bibitem[Charalampakis et~al., 2016]{charalampakis2016comparison}
Charalampakis, B., Spathis, D., Kouslis, E., and Kermanidis, K. (2016).
\newblock A comparison between semi-supervised and supervised text mining techniques on detecting irony in greek political tweets.
\newblock {\em Engineering Applications of Artificial Intelligence}, 51:50--57.

\bibitem[Davidson et~al., 2017]{davidson2017automated}
Davidson, T., Warmsley, D., Macy, M., and Weber, I. (2017).
\newblock Automated hate speech detection and the problem of offensive language.
\newblock In {\em Eleventh international aaai conference on web and social media}.

\bibitem[Del~Vigna et~al., 2017]{del2017hate}
Del~Vigna, F., Cimino, A., Dell’Orletta, F., Petrocchi, M., and Tesconi, M. (2017).
\newblock Hate me, hate me not: Hate speech detection on facebook.
\newblock {\em Proceedings of the First Italian Conference on Cybersecurity (ITASEC17), Venice, Italy}.

\bibitem[Di~F{\'a}tima et~al., 2023]{di2023hate}
Di~F{\'a}tima, B., Munoriyarwa, A., Gilliland, A., Msughter, A.~E., Vizca{\'\i}no-Verd{\'u}, A., G{\"o}kaliler, E., Capoano, E., Yu, H., Alik{\i}l{\i}{\c{c}}, {\.I}., Gonz{\'a}lez-Aguilar, J.-M., et~al. (2023).
\newblock {\em Hate Speech on Social Media: A Global Approach}.
\newblock Pontificia Universidad Cat{\'o}lica del Ecuador.

\bibitem[DistilBERT base model (uncased), 2024]{DistilBERT}
DistilBERT base model (uncased) (2024).
\newblock https://huggingface.co/distilbert/distilbert-base-uncased.

\bibitem[D{\"o}ring and Mohseni, 2019]{doring2019fail}
D{\"o}ring, N. and Mohseni, M.~R. (2019).
\newblock Fail videos and related video comments on youtube: a case of sexualization of women and gendered hate speech?
\newblock {\em Communication research reports}, 36(3):254--264.

\bibitem[D{\"o}ring and Mohseni, 2020]{doring2020gendered}
D{\"o}ring, N. and Mohseni, M.~R. (2020).
\newblock Gendered hate speech in youtube and younow comments: Results of two content analyses.
\newblock {\em SCM Studies in Communication and Media}, 9(1):62--88.

\bibitem[Europe, 2014]{ilga}
Europe, I. (2014).
\newblock {Hate crime \& hate speech} retrieved from.

\bibitem[{European Commission}, 1999]{euc}
{European Commission} (1999).
\newblock {Speech by Commissioner Jourová - 10 years of the EU Fundamental Rights Agency: a call to action in defence of fundamental rights, democracy and the rule of law} retrieved from.

\bibitem[Facebook, 2024]{facebook}
Facebook (2024).
\newblock How do i report inappropriate or abusive things on facebook.
\newblock https://www.facebook.com/help/212722115425932.

\bibitem[Fortuna and Nunes, 2018]{fortuna2018survey}
Fortuna, P. and Nunes, S. (2018).
\newblock A survey on automatic detection of hate speech in text.
\newblock {\em ACM Computing Surveys (CSUR)}, 51(4):85.

\bibitem[Gaffney, 2018]{gab2020pushshift}
Gaffney, D. (2018).
\newblock The pushshift {Gab} dataset.

\bibitem[Jahan and Oussalah, 2023]{jahan2023systematic}
Jahan, M.~S. and Oussalah, M. (2023).
\newblock A systematic review of hate speech automatic detection using natural language processing.
\newblock {\em Neurocomputing}, page 126232.

\bibitem[Joulin et~al., 2016]{joulin2016fasttext}
Joulin, A., Grave, E., Bojanowski, P., Douze, M., J{\'e}gou, H., and Mikolov, T. (2016).
\newblock Fasttext.zip: Compressing text classification models.
\newblock {\em arXiv preprint arXiv:1612.03651}.

\bibitem[Kazemi et~al., 2022]{kazemi2022research}
Kazemi, A., Garimella, K., Shahi, G.~K., Gaffney, D., and Hale, S.~A. (2022).
\newblock Research note: Tiplines to uncover misinformation on encrypted platforms: A case study of the 2019 indian general election on whatsapp.
\newblock {\em Harvard Kennedy School Misinformation Review}.

\bibitem[Li et~al., 2023]{li2023hot}
Li, L., Fan, L., Atreja, S., and Hemphill, L. (2023).
\newblock "hot" {ChatGPT}: The promise of chatgpt in detecting and discriminating hateful, offensive, and toxic comments on social media.
\newblock {\em arXiv preprint arXiv:2304.10619}.

\bibitem[Liu et~al., 2019]{liu2019nuli}
Liu, P., Li, W., and Zou, L. (2019).
\newblock Nuli at semeval-2019 task 6: transfer learning for offensive language detection using bidirectional transformers.
\newblock In {\em Proceedings of the 13th International Workshop on Semantic Evaluation}, pages 87--91.

\bibitem[Loper and Bird, 2002]{loper2002nltk}
Loper, E. and Bird, S. (2002).
\newblock Nltk: the natural language toolkit.
\newblock {\em arXiv preprint cs/0205028}.

\bibitem[Malmasi and Zampieri, 2017]{malmasi2017detecting}
Malmasi, S. and Zampieri, M. (2017).
\newblock Detecting hate speech in social media.
\newblock In {\em Proceedings of the International Conference Recent Advances in Natural Language Processing, RANLP 2017}, pages 467--472.

\bibitem[Mandl et~al., 2019]{mandl2019overview}
Mandl, T., Modha, S., Majumder, P., Patel, D., Dave, M., Mandlia, C., and Patel, A. (2019).
\newblock Overview of the hasoc track at fire 2019: Hate speech and offensive content identification in indo-european languages.
\newblock In {\em Proceedings of the 11th Forum for Information Retrieval Evaluation}, pages 14--17.

\bibitem[McHugh, 2012]{mchugh2012interrater}
McHugh, M.~L. (2012).
\newblock Interrater reliability: the kappa statistic.
\newblock {\em Biochemia medica}, 22(3):276--282.

\bibitem[Nobata et~al., 2016]{nobata2016abusive}
Nobata, C., Tetreault, J., Thomas, A., Mehdad, Y., and Chang, Y. (2016).
\newblock Abusive language detection in online user content.
\newblock In {\em Proceedings of the 25th international conference on world wide web}, pages 145--153.

\bibitem[Nockleby, 2000]{nockleby2000hate}
Nockleby, J.~T. (2000).
\newblock Hate speech.
\newblock {\em Encyclopedia of the American constitution}, 3:1277--1279.

\bibitem[Ousidhoum et~al., 2019]{ousidhoum2019multilingual}
Ousidhoum, N., Lin, Z., Zhang, H., Song, Y., and Yeung, D.-Y. (2019).
\newblock Multilingual and multi-aspect hate speech analysis.
\newblock In {\em Proceedings of the 2019 Conference on Empirical Methods in Natural Language Processing and the 9th International Joint Conference on Natural Language Processing (EMNLP-IJCNLP)}, pages 4675--4684.

\bibitem[Pedregosa et~al., 2011]{scikit-learn}
Pedregosa, F., Varoquaux, G., Gramfort, A., Michel, V., Thirion, B., Grisel, O., Blondel, M., Prettenhofer, P., Weiss, R., Dubourg, V., Vanderplas, J., Passos, A., Cournapeau, D., Brucher, M., Perrot, M., and Duchesnay, E. (2011).
\newblock Scikit-learn: Machine learning in {P}ython.
\newblock {\em Journal of Machine Learning Research}, 12:2825--2830.

\bibitem[Poletto et~al., 2021]{poletto2021resources}
Poletto, F., Basile, V., Sanguinetti, M., Bosco, C., and Patti, V. (2021).
\newblock Resources and benchmark corpora for hate speech detection: a systematic review.
\newblock {\em Language Resources and Evaluation}, 55:477--523.

\bibitem[Prolific, 2024]{prolific}
Prolific (2024).
\newblock www.prolific.com.

\bibitem[Reimers and Gurevych, 2020]{reimers2020making}
Reimers, N. and Gurevych, I. (2020).
\newblock Making monolingual sentence embeddings multilingual using knowledge distillation.
\newblock In {\em Proceedings of the 2020 Conference on Empirical Methods in Natural Language Processing (EMNLP)}, pages 4512--4525.

\bibitem[R{\"o}chert et~al., 2020]{rochert2020opinion}
R{\"o}chert, D., Neubaum, G., Ross, B., Brachten, F., and Stieglitz, S. (2020).
\newblock Opinion-based homogeneity on youtube.
\newblock {\em Computational Communication Research}, 2(1):81--108.

\bibitem[R{\"o}chert et~al., 2021]{rochert2021networked}
R{\"o}chert, D., Shahi, G.~K., Neubaum, G., Ross, B., and Stieglitz, S. (2021).
\newblock The networked context of covid-19 misinformation: informational homogeneity on youtube at the beginning of the pandemic.
\newblock {\em Online Social Networks and Media}, 26:100164.

\bibitem[Ross et~al., 2016]{ross2016hatespeech}
Ross, B., Rist, M., Carbonell, G., Cabrera, B., Kurowsky, N., and Wojatzki, M. (2016).
\newblock {Measuring the Reliability of Hate Speech Annotations: The Case of the European Refugee Crisis}.
\newblock In Bei{\ss}wenger, M., Wojatzki, M., and Zesch, T., editors, {\em Proceedings of NLP4CMC III: 3rd Workshop on Natural Language Processing for Computer-Mediated Communication}, volume~17 of {\em Bochumer Linguistische Arbeitsberichte}, pages 6--9, Bochum.

\bibitem[Salminen et~al., 2020]{salminen2020developing}
Salminen, J., Hopf, M., Chowdhury, S.~A., Jung, S.-g., Almerekhi, H., and Jansen, B.~J. (2020).
\newblock Developing an online hate classifier for multiple social media platforms.
\newblock {\em Human-centric Computing and Information Sciences}, 10(1):1.

\bibitem[Shahi and Kana~Tsoplefack, 2022]{shahi2022mitigating}
Shahi, G.~K. and Kana~Tsoplefack, W. (2022).
\newblock Mitigating harmful content on social media using an interactive user interface.
\newblock In {\em Social Informatics: 13th International Conference, SocInfo 2022, Glasgow, UK, Proceedings}, pages 490--505. Springer.

\bibitem[Shahi and Majchrzak, 2022]{shahi2022amused}
Shahi, G.~K. and Majchrzak, T.~A. (2022).
\newblock Amused: an annotation framework of multimodal social media data.
\newblock In {\em Intelligent Technologies and Applications: 4th International Conference, INTAP 2021, Grimstad, Norway, October 11--13, 2021, Revised Selected Papers}, pages 287--299. Springer.

\bibitem[Siegel, 2020]{siegel2020online}
Siegel, A.~A. (2020).
\newblock Online hate speech.
\newblock {\em Social media and democracy: The state of the field, prospects for reform}, pages 56--88.

\bibitem[Stoll et~al., 2020]{stoll2020detecting}
Stoll, A., Ziegele, M., and Quiring, O. (2020).
\newblock Detecting incivility and impoliteness in online discussions.
\newblock {\em Computational Communication Research}, 2(1):109--134.

\bibitem[Wang and Kim, 2023]{wang2023content}
Wang, S. and Kim, K.~J. (2023).
\newblock Content moderation on social media: does it matter who and why moderates hate speech?
\newblock {\em Cyberpsychology, Behavior, and Social Networking}, 26(7):527--534.

\bibitem[Waseem, 2016]{waseem2016you}
Waseem, Z. (2016).
\newblock Are you a racist or am i seeing things? annotator influence on hate speech detection on twitter.
\newblock In {\em Proceedings of the first workshop on NLP and computational social science}, pages 138--142.

\bibitem[Waseem and Hovy, 2016]{waseem2016hateful}
Waseem, Z. and Hovy, D. (2016).
\newblock Hateful symbols or hateful people? predictive features for hate speech detection on twitter.
\newblock In {\em Proceedings of the NAACL student research workshop}, pages 88--93.

\bibitem[Wei et~al., 2017]{wei2017convolution}
Wei, X., Lin, H., Yang, L., and Yu, Y. (2017).
\newblock A convolution-lstm-based deep neural network for cross-domain mooc forum post classification.
\newblock {\em Information}, 8(3):92.

\bibitem[Wigand and Voin, 2017]{wigand2017speech}
Wigand, C. and Voin, M. (2017).
\newblock Speech by commissioner jourov{\'a}—10 years of the eu fundamental rights agency: A call to action in defence of fundamental rights, democracy and the rule of law.

\bibitem[Wullach et~al., 2020]{wullach2020towards}
Wullach, T., Adler, A., and Minkov, E. (2020).
\newblock Towards hate speech detection at large via deep generative modeling.
\newblock {\em IEEE Internet Computing}, 25(2):48--57.

\bibitem[X.com, 2024]{twitter}
X.com (2024).
\newblock The twitter rules.
\newblock https://help.twitter.com/en/rules-and-policies/x-rules.

\bibitem[YouTube, 2024]{youtube}
YouTube (2024).
\newblock Hate speech policy: Youtube community guidelines.

\bibitem[Yuan et~al., 2023]{yuan2023transfer}
Yuan, L., Wang, T., Ferraro, G., Suominen, H., and Rizoiu, M.-A. (2023).
\newblock Transfer learning for hate speech detection in social media.
\newblock {\em Journal of Computational Social Science}, 6(2):1081--1101.

\bibitem[Zampieri et~al., 2019]{zampieri2019semeval}
Zampieri, M., Malmasi, S., Nakov, P., Rosenthal, S., Farra, N., and Kumar, R. (2019).
\newblock Semeval-2019 task 6: Identifying and categorizing offensive language in social media (offenseval).
\newblock In {\em Proceedings of the 13th International Workshop on Semantic Evaluation}, pages 75--86.

\end{thebibliography}

\end{document}